\documentclass{article}

\usepackage{microtype}
\usepackage{graphicx}
\usepackage{booktabs} 
\usepackage{caption}
\usepackage{subcaption}
\usepackage{wrapfig}
\usepackage{multirow}
\usepackage{forest}
\usepackage[shortlabels]{enumitem}
\usepackage{url}

\usepackage{hyperref}

\usepackage[ruled, vlined, linesnumbered,algosection]{algorithm2e}

\usepackage[accepted]{icml2024}

\usepackage{amsmath}
\usepackage{amssymb}
\usepackage{mathtools}
\usepackage{amsthm}
\usepackage{cancel}
\usepackage{siunitx}
\usepackage{amsmath,amsfonts,bm}

\def\eqref#1{equation~\ref{#1}}

\def\1{\bm{1}}

\DeclareMathAlphabet{\mathsfit}{\encodingdefault}{\sfdefault}{m}{sl}
\SetMathAlphabet{\mathsfit}{bold}{\encodingdefault}{\sfdefault}{bx}{n}

\newcommand{\E}{\mathbb{E}}

\newcommand{\R}{\mathbb{R}}

\usepackage[capitalize,noabbrev]{cleveref}

\defcitealias{united_nations_office_for_disaster_risk_reduction_undrr_human_2015}{United Nations, 2015}
\defcitealias{united_nations_office_for_disaster_risk_reduction_undrr_human_2019}{UNDRR \& CRED, 2019}
\defcitealias{united_nations_office_for_disaster_risk_reduction_undrr_global_2022}{United Nations, 2022}
\defcitealias{centre_for_research_on_the_epidemiology_of_disasters_cred_disasters_2022}{CRED, 2022}

\Crefname{algocf}{Algorithm}{Algorithms}

\SetCommentSty{commentfont}
\let\oldnl\nl
\newcommand{\nonl}{\renewcommand{\nl}{\let\nl\oldnl}}

\theoremstyle{plain}

\theoremstyle{definition}

\theoremstyle{remark}

\newcommand{\mat}[1]{\mathbf{#1}}
\newcommand{\vect}[1]{\bm{#1}}

\renewcommand{\R}{\mathbb{R}}

\DeclareMathOperator{\score}{\varrho}

\newcommand{\graph}{\mathcal{G}}
\newcommand{\vertices}{\mathcal{V}}
\newcommand{\edges}{\mathcal{E}}
\newcommand{\predecessors}{\vertices_{\mathrm{in}}}
\newcommand{\successors}{\vertices_{\mathrm{out}}}
\newcommand{\featMat}{\mat{X}}

\newcommand{\trueVecSymb}{y}
\newcommand{\trueVec}{\vect{\trueVecSymb}}
\newcommand{\predVecSymb}{\hat{\trueVecSymb}}
\newcommand{\predVec}{\vect{\predVecSymb}}
\newcommand{\qMatSymb}{Q}
\newcommand{\qMat}{\mat{\qMatSymb}}
\newcommand{\qVec}{\bm{q}}
\newcommand{\adjMatSymb}{A}
\newcommand{\adjMat}{\mat{\adjMatSymb}}

\newcommand{\adjMatNormSymb}{\bar{\adjMatSymb}}
\newcommand{\adjMatNorm}{\mat{\adjMatNormSymb}}
\newcommand{\dIn}{\mathrm{in}}

\newcommand{\degMatSymb}{D}
\newcommand{\degMat}{\mat{\degMatSymb}}

\DeclarePairedDelimiter{\abs}{\lvert}{\rvert}
\DeclarePairedDelimiter{\norm}{\lVert}{\rVert}

\DeclareMathOperator{\diag}{diag}
\newcommand{\idMat}{\mathbf{I}}

\newcommand{\intMixMatSymb}{\Theta}
\newcommand{\intMixMat}{\mat{\intMixMatSymb}}

\newcommand{\act}{\sigma}

\newcommand{\Params}{\bm{\Theta}}
\DeclareMathOperator{\ReLU}{ReLU}

\DeclareMathOperator{\ResGCNConv}{ResGCNLayer}
\DeclareMathOperator{\GCNIIConv}{GCNIILayer}
\DeclareMathOperator{\ResGATLayer}{ResGATLayer}

\newcommand{\gnnlayer}{\operatorname{GNNLayer}}
\newcommand{\encoder}{\operatorname{Encoder}}
\newcommand{\decoder}{\operatorname{Decoder}}

\renewcommand{\E}{\mathbb{E}}

\newcommand{\meanVec}{\bm{\mu}}
\newcommand{\stdVec}{\bm{\sigma}}

\DeclareMathOperator{\nse}{NSE}

\newcommand{\cell}[2]{\begin{tabular}[x]{@{}r@{}}#1\\\textcolor{gray}{\tiny $\pm$#2}\end{tabular}}

\DeclareFontFamily{U}{matha}{\hyphenchar\font45}
\DeclareFontShape{U}{matha}{m}{n}{
	<5> <6> <7> <8> <9> <10> gen * matha
	<10.95> matha10 <12> <14.4> <17.28> <20.74> <24.88> matha12
}{}
\DeclareSymbolFont{matha}{U}{matha}{m}{n}
\DeclareMathSymbol{\odiv}         {2}{matha}{"63}

\makeatletter
\def\moverlay{\mathpalette\mov@rlay}
\def\mov@rlay#1#2{\leavevmode\vtop{%
		\baselineskip\z@skip \lineskiplimit-\maxdimen
		\ialign{\hfil$\m@th#1##$\hfil\cr#2\crcr}}}
\newcommand{\charfusion}[3][\mathord]{
	#1{\ifx#1\mathop\vphantom{#2}\fi
		\mathpalette\mov@rlay{#2\cr#3}
	}
	\ifx#1\mathop\expandafter\displaylimits\fi}
\makeatother

\newcommand{\cupdot}{\charfusion[\mathbin]{\cup}{\cdot}}

\usepackage[textsize=tiny]{todonotes}  

\makeatletter
\define@key{todonotes}{N}[]{%
    \setkeys{todonotes}{author=\textbf{Nikolas},color=blue!50!white}}%
\define@key{todonotes}{Y}[]{%
    \setkeys{todonotes}{author=\textbf{Yixuan},color=green!50!white}}%
\makeatother

\icmltitlerunning{The Merit of River Network Topology for Neural Flood Forecasting}

\begin{document}

\twocolumn[
\icmltitle{The Merit of River Network Topology for Neural Flood Forecasting}



\icmlsetsymbol{equal}{*}

\begin{icmlauthorlist}
\icmlauthor{Nikolas Kirschstein}{ox,tum}
\icmlauthor{Yixuan Sun}{ox}
\end{icmlauthorlist}

\icmlaffiliation{ox}{Mathematical Institute, University of Oxford, UK \vfill}
\icmlaffiliation{tum}{Department of Informatics, Technical University of Munich, Germany}

\icmlcorrespondingauthor{Nikolas Kirschstein\\}{{\url{nikolas.kirschstein@maths.ox.ac.uk}}}

\icmlkeywords{Machine Learning, ICML}

\vskip 0.3in
]



\printAffiliationsAndNotice{}  
\setcounter{footnote}{0}

\begin{abstract}
Climate change exacerbates riverine floods, which occur with higher frequency and intensity than ever. The much-needed forecasting systems typically rely on accurate river discharge predictions. To this end, the SOTA data-driven approaches treat forecasting at spatially distributed gauge stations as isolated problems, even within the same river network. However, incorporating the known topology of the river network into the prediction model has the potential to leverage the adjacency relationship between gauges. Thus, we model river discharge for a network of gauging stations with GNNs and compare the forecasting performance achieved by different adjacency definitions.
Our results show that the model fails to benefit from the river network topology information, both on the entire network and small subgraphs. The learned edge weights correlate with neither of the static definitions and exhibit no regular pattern. Furthermore, the GNNs struggle to predict sudden, narrow discharge spikes. Our work hints at a more general underlying phenomenon of neural prediction not always benefitting from graphical structure and may inspire a systematic study of the conditions under which this happens.
\end{abstract}
\section{Introduction}

\begin{figure}
    \centering
    \hspace*{-2mm}
	\includegraphics[width=0.49\textwidth]{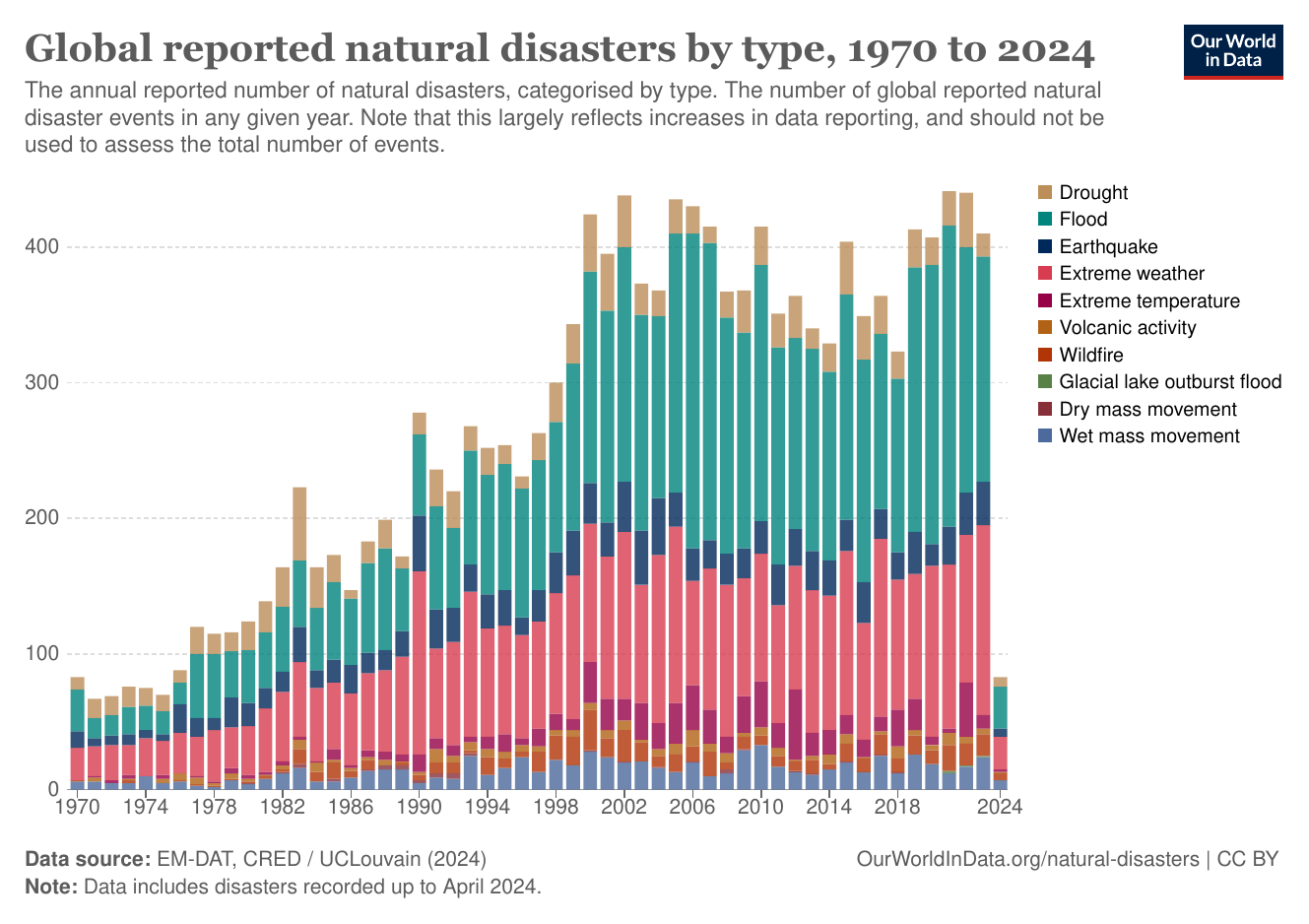}
	\caption{Historical occurrence of natural disasters by disaster type. The number of events increased over time, with floods being the most common. \citep{ritchie_natural_2022}.}
	\label{fig:disasters}
 \vspace*{-3mm}
\end{figure}

Floods are among the most destructive natural disasters that occur on Earth, causing extensive damage to infrastructure, property, and human life. They are also the most common type of disaster, accounting for almost half of all disaster events recorded (cp. \cref{fig:disasters}).
In 2022 alone, floods affected 57.1 million people worldwide, killed almost 8000, and caused 44.9 billion USD in damages \citepalias{centre_for_research_on_the_epidemiology_of_disasters_cred_disasters_2022}.
With climate change ongoing, floods have become increasingly frequent over the last decades and are expected to be even more prevalent in the future \citepalias{united_nations_office_for_disaster_risk_reduction_undrr_global_2022}. Thus, early warning systems that can help authorities and individuals prepare for and respond to impending floods play a crucial role in mitigating fatalities and economic costs.

Operational forecasting systems such as Google's Flood Forecasting Initiative \citep{nevo_flood_2022} typically focus on riverine floods, which are responsible for the vast majority of damages. A key component in these systems is the prediction of future river discharge\footnote{amount of water volume passing through a given river section per unit time} at a gauging station based on environmental indicators such as past discharge and precipitation. The state-of-the-art data-driven approaches are based on \citet{kratzert_towards_2019} and consist in training an LSTM variant on multiple gauges jointly  to exploit the shared underlying physics. However, even when some of those gauges are in the same river network, this topology information is not taken into account. One reason might be that the main benchmarking dataset family CAMELS-x \citep{addor_camels_2017,alvarez-garreton_camels-cl_2018,coxon_camels-gb_2020,chagas_camels-br_2020,fowler_camels-aus_2021} does not contain such information. Recently, \citet{klingler_lamah-ce_2021} published a new benchmarking dataset LamaH-CE that follows the CAMELS-x framework but includes topology data.

In this work, we investigate the effect of river network topology information on discharge predictions by employing a single end-to-end GNN to allow the network structure to be utilised during the prediction process. We train GNNs on LamaH-CE and, to assess the merit of incorporating the graph structure, compare the effect of different adjacency definitions: \vspace*{-0.5mm}
\begin{enumerate}[(1)]
	\item
	no adjacency, which is equivalent to existing approaches with cross-gauge shared parameters but isolated gauges,
	\vspace*{-0.5mm}
	\item 
	binary adjacency of neighbouring gauges in the network,
	\vspace*{-0.5mm}
	\item 
	weighted adjacency according to physical relationships, namely stream length, elevation difference, and average slope between neighbouring gauges, and
	\vspace*{-0.5mm}
	\item 
	learned adjacency by treating edge weights as a model parameter.
\end{enumerate}

\vspace*{-0.5mm}
We perform this comparison for both the entire dataset as well as four deliberately chosen small-scale subnetworks with different local topologies. Furthermore, we inspect how the learned edge weights in (4) correlate with the static weights in (3).  Finally, we analyse the model's behaviour on the worst-performing gauge. Our source code is publicly available at \url{https://github.com/nkirschi/neural-flood-forecasting}.
\section{Related Work}

Classical approaches towards river discharge prediction stem from finite-element solutions to partial differential equations such as the Saint-Venant shallow-water equations \citep{vreugdenhil_numerical_1994,wu_computational_2007}. However, these models suffer from scalability issues since they become computationally prohibitive on larger scales, as required in the real world \citep{nevo_ml-based_2020}. Furthermore, they impose a strong inductive bias by making numerous assumptions about the underlying physics.

On the other hand, data-driven methods and in particular deep learning provide excellent scaling properties and are less inductively biased. They are increasingly being explored for a plethora of hydrological applications, including discharge prediction \citep[see surveys by][]{mosavi_flood_2018,chang_flood_2019,sit_comprehensive_2020}, where they tend to achieve higher accuracy than the classical models. The vast majority of studies employ Long Short-Term Memory models \citep[LSTM;][]{hochreiter_long_1997} due to their inherent suitability for sequential tasks and reliability in predicting extreme events \citep{frame_deep_2022}. Whereas these studies usually consider forecasting for a single gauging station, \citet{kratzert_toward_2019,kratzert_towards_2019} demonstrate the generalisation benefit of training a single spatially distributed LSTM model on multiple gauging sites jointly. Their approach exploits the shared underlying physics across gauges but is still agnostic to the relationship between sites. 

Incorporating information from neighbouring stations or even an entire river network into a spatially distributed model potentially improves prediction performance.  Upstream gauges could ``announce" the advent of  increased water masses to downstream gauges, which in turn could provide forewarning about already ongoing flooding further downstream. The input then becomes a graph whose vertices represent gauges and edges represent flow between gauges. The corresponding deep learning tool to capture these spatial dependencies is Graph Neural Networks (GNN). \citet{kratzert_large-scale_2021} employ it as a post-processing step to route the per-gauge discharge predicted by a conventional LSTM along the river network. In contrast, we seek to unify prediction and routing in a single GNN.
\vspace*{-1mm}
\section{Methodology}

\subsection{Data Preprocessing} \label{sec:preprocessing}

\begin{figure}[t]
	\centering
	\includegraphics[width=\linewidth]{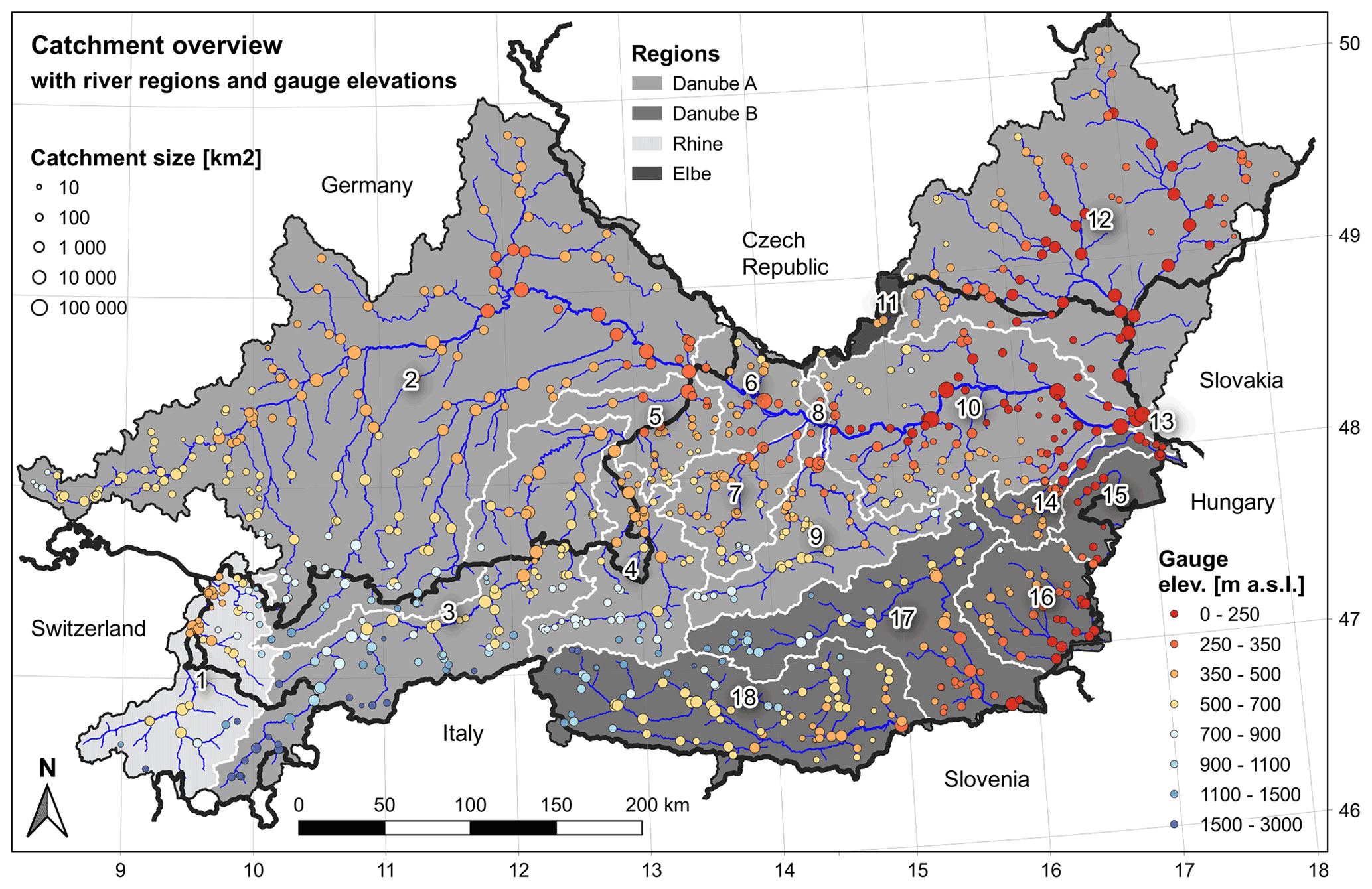}
	\caption{Geographical contextualisation of the LamaH-CE dataset. Circle colour indicates gauge elevation, while circle size indicates catchment size. \citep{klingler_lamah-ce_2021}}
	\label{fig:map}
 \vspace*{-5mm}
\end{figure}

The LamaH-CE\footnote{\textbf{LA}rge-Sa\textbf{M}ple D\textbf{A}ta for \textbf{H}ydrology for \textbf{C}entral \textbf{E}urope} dataset \citep{klingler_lamah-ce_2021} contains historical discharge and meteorological measurements on an hourly resolution for 859 gauges in the broader Danube river network shown in \cref{fig:map}. Covering an area of \qty{170000}{km^2} with diverse environmental conditions, \citeauthor{klingler_lamah-ce_2021} expect that results from investigations on this dataset carry over to other river networks. 
Unfortunately, LamaH-CE does not provide any flood event annotations, so that we can only model continuous discharge but not floods as discrete events. Moreover, the dataset does not include average propagation time between gauges, meaning that a predictor needs to implicitly infer the time lag by comparing observations at neighbouring gauges.
\newpage
The river network defined by LamaH-CE naturally forms a directed acyclic graph (DAG) $\graph = (\vertices, \edges)$. The nodes $\vertices$ represent gauges, and the edges $\edges$ represent flow between a gauge and the next downstream gauges. Hence, $\graph$ is \emph{anti-transitive}, i.e., no skip connections exist. We preprocess $\graph$ to distil a connected subgraph with complete data.

\vspace{2mm}

\textbf{Region Selection.} \cref{fig:map} shows that $\graph$ contains four different connected components, of which we restrict ourselves to the largest one, "Danube A". Its most downstream gauge close to the Austrian-Hungarian border has complete discharge data for the years 2000 through 2017. Starting at this gauge, we determine all connected gauges of the Danube A region by performing an inverse depth-first search given by \cref{algo:invserseDFS}. Overall, 608 out of the original 859 gauges belong to this connected component.

\textbf{Gauge Filtering.} While the meteorological data is complete, the discharge data contains gaps. \citeauthor{klingler_lamah-ce_2021} have filled any consecutive gaps of at most six hours by linear interpolation and left the remaining longer gaps unaltered. We only want to consider gauges that (a) do not have these longer periods of missing values and (b) provide discharge data for at least the same time frame (2000 to 2017) as the most downstream gauge. To this end, we remove all gauges that violate these requirements from the graph using \cref{algo:softRemoval}. Predecessors and successors of a deleted node get newly connected with a combined edge weight so that network connectivity is maintained. Note that thanks to antitransitivity, a duplicate check is unnecessary when inserting the new edges. After this preprocessing step, we are left with 358 out of the previously 608 gauges.

\vspace{2mm}

Overall, the reduced graph $\graph$ consists of $n \coloneqq \abs{\vertices} = 358$ gauges with $T$ hours of discharge measurements for the years 2000 to 2017, which we can conceptually represent as a node signal $\qMat = \begin{bmatrix}\qVec^{(1)} \;\vert\; \qVec^{(2)} \;\vert\; \dots \;\vert\; \qVec^{(T)}\end{bmatrix} \in \R^{n \times T}$. Equally, we have four node signals of the same shape for each of the meteorological context variables precipitation, topsoil moisture, air temperature, and surface pressure. However, we exclude them from all notation throughout the paper for simplicity of presentation, i.e., drop the implicit third dimension of size $5$.

\vspace{2mm}

\textbf{Normalisation.}  We normalise the data to surrender all gauges to the same scale and accelerate the training process \citep{lecun_efficient_2002}. In particular, we normalise per gauge, i.e., element-wise, using the standard score:
\[
\qVec^{(t)} \gets \frac{\qVec^{(t)} - \meanVec}{\stdVec} \quad \text{ where } \begin{array}{r@{\hphantom{.}}l}
    \meanVec &= \frac{1}{T}\sum_{t = 1}^{T} \qVec^{(t)} \\
    \stdVec^2 &= \frac{1}{T - 1}\sum_{i = 1}^{T}(\qVec^{(t)} - \meanVec)^2
\end{array}
\]
\textbf{Train-test splits.} To robustly assess the performance of a trained model on unseen data via cross-validation, we consider three different train-test splits. The last two years 2016 and 2017 always serve for testing, and from the remainder eight years are chosen for training: once the even years in 2000 to 2015, once the odd years in 2000 to 2015, and once the contiguous years 2008 to 2015. Note the differences to vanilla fold-based cross-validation schemes: (a) we need to ensure the train years temporally precede the test years, and (b) due to the small amount of available years we choose the same test set for all splits.

\subsection{The Forecasting Task}
\label{sec:graphdef}

We task the model with an instance of supervised node regression. Assume we are given a certain amount of $W$ (\emph{"window size"}) most recent hours of discharge and meteorological measurements, for all gauges. Our goal is to predict the discharge $L$ (\emph{"lead time"}) hours in the future. Again, for simplicity, we restrict all notation to the discharge data in the input since the meteorological data can be trivially added in an extra dimension.

\textbf{Features \& Targets.} To conduct supervised learning, we extract input-output pairs from the time series represented by $\qMat$ (cp. \cref{sec:preprocessing}). For $t = W, W+1 \dots, T - L$, we define the feature matrix at time step $t$ as
\[
\featMat^{(t)} \coloneqq \begin{bmatrix}
	\qVec^{(t-W+1)} \;\Bigg\vert\; \dots \;\Bigg\vert\; \qVec^{(t-1)} \;\Bigg\vert\; \qVec^{(t)}
\end{bmatrix} \in \R^{n \times W}
\]
and the corresponding target vector as
$\trueVec^{(t)} \coloneqq \qVec^{(t+L)} \in \R^{n}$.
We collect all samples into the set $\mathcal{D} = \{(\featMat^{(t)}, \trueVec^{(t)})\}_{t=W}^{T-L}$ and partition it according to a given train-test split into $\mathcal{D} = \mathcal{D}_\text{train} \cupdot \mathcal{D}_\text{test}$. 
The extraction process can be illustrated as follows:
\[
\hspace*{-2mm}
\begin{array}{ll}
	& \xlongrightarrow{\hspace{36mm} \text{time} \hspace{36mm}}\vspace*{-1mm} \\
	\raisebox{-12mm}{\rotatebox{90}{$\xlongleftarrow{\hspace*{8mm} \text{gauges} \hspace*{8mm}}$}}\hspace*{-4mm} &
	\begin{bmatrix}
		\hspace*{7mm}\raisebox{5mm}{\textcolor{gray}{$\mathbf{Q}$}}\hspace*{7mm} & \fbox{$\begin{array}{c}\vspace*{4mm}\\\hspace*{5mm}\featMat^{(t)}\hspace*{3mm}\\\vspace*{4mm}\end{array}$} \hspace*{-1mm}\xleftrightarrow{\hspace*{2mm}L\hspace*{2mm}}\hspace*{-1mm} \fbox{$\begin{array}{c}\vspace*{4mm}\\\hspace*{-3mm}\trueVec^{(t)}\hspace*{-3.5mm}\\\vspace*{4mm}\end{array}$} & \hspace*{5mm}\dots \hspace*{5mm}
	\end{bmatrix} \vspace*{-3mm}\\
	& \hspace*{23mm}\underbrace{\hspace*{20mm}}_{W}
\end{array}
\]

\textbf{Adjacency.} Besides the input and target measurements, we feed the river network topology to the GNN in the form of an adjacency matrix $\adjMat \in \R^{n \times n}$. For the definition of matrix entries corresponding to an edge $(i,j) \in \edges$ (the rest being zero), we consider the following choices:
\begin{enumerate}[(1)]
	\item \emph{isolated:} $\adjMat_{i,j} \coloneqq 0$ equates to removing all edges and results in the augmented normalised adjacency matrix to be a multiple of the identity so that each GNN layer degenerates to a node-wise linear layer.
	\item \emph{binary:} $\adjMat_{i,j} \coloneqq 1$ corresponds to the unaltered adjacency matrix as it comes with the LamaH-CE dataset.
 \newpage
	\item \emph{weighted:} $\adjMat_{i,j} \coloneqq {w}_{(i,j)}$ quantifies a physical relationship, for which LamaH-CE provides three alternatives:
	\begin{itemize}
	\item the \emph{stream length} along the river between $i$ and $j$,
	\item the \emph{elevation difference} along the river between $i$ and $j$, and
	\item the \emph{average slope} of the river between $i$ and $j$.
	\end{itemize}
	\item \emph{learned:} $\adjMat_{i,j} \coloneqq {\omega}_{(i,j)}$ where $\vect{\omega} \in \R^{\abs{\edges}}$ is a learnable model parameter.
\end{enumerate}
The first two variants allow us to compare the effect of introducing the river network topology into the model at all. The last two variants enable insights into what kind of relative importance of edges is most helpful. As usual in GNNs, we use the normalised augmented adjacency matrix 
\[
\adjMatNorm \coloneqq (\degMat_\dIn + \diag(\bm{\xi}))^{-\frac{1}{2}}(\adjMat  + \diag(\bm{\xi}))(\degMat_\dIn + \diag(\bm{\xi}))^{-\frac{1}{2}}
\]
where self-loops  for node $i$ with weight $\xi_i$ are added and everything is symmetrically normalised based on the diagonal in-degree matrix $\degMat_\dIn$. We generally set $\xi_i$ as the mean of all incoming edge weights at node $i$ to make self-loops roughly equally important to the other edges. The only exception to this is option (1) above, where that mean would be zero and thus result in no information flow whatsoever, so that in this case, we set the self-loop weights to one instead.

\textbf{Model.} Our desideratum is a GNN $f_\theta: \R^{n \times W} \to \R^n$ parameterised by $\theta$ which closely approximates the mapping of windows $\featMat$ to targets $\trueVec$, i.e., $\predVec \coloneqq f_\theta(\featMat) \approx \trueVec$. All our models have a sandwich architecture: an affine layer $\encoder_{\Params_0}: \R^{n \times W} \to \R^{n \times d}$ embeds the $W$-dimensional input per gauge into a $d$-dimensional latent space. On this space, a sequence of $N$ layers $\gnnlayer_{\Params_i}: \R^{n \times d} \times \R^{n \times n} \to \R^{n \times d}$ with subsequent activation function $\act = \ReLU$ are applied. Finally, another affine layer $\decoder_{\Params_{N+1}}: \R^{n \times d} \to \R^n$ projects from the latent space to a scalar per gauge. In symbols:
\begin{equation*}
\begin{split}
\mathbf{H}^{(0)} &\coloneqq \encoder_{\Params_0}(\featMat) \\
\mathbf{H}^{(i)} &\coloneqq \sigma(\gnnlayer_{\Params_i}(\mathbf{H}^{(i-1)}, \adjMatNorm) )\quad \text{for } i = 1, \dots, N\\
\predVec &\coloneqq \decoder_{\Params_{N+1}}(\mathbf{H}^{(N)}).
\end{split}
\end{equation*}
We consider three choices for $\gnnlayer$: a residual version of the vanilla GCN layer \citep{kipf_semi-supervised_2017}, the inherently residual GCNII layer \citep{chen_simple_2020}, and a residual version of the attention-based GAT layer \citep{velickovic_graph_2017}. Since the GAT layer already contains a learned component, the adjacency case (4) would be redundant for this architecture, so that we replace it with the case of providing all three edge weights in (3) jointly, which is not possible with the other two layer definitions. All three employ residual connections to overcome the phenomenon known as \emph{oversmoothing} \citep{oono_graph_2020}, where the features of adjacent nodes converge with increasing depth.

\begin{figure*}[t]
\label{fig:relevancy}
    \centering
    \includegraphics[width=\linewidth]{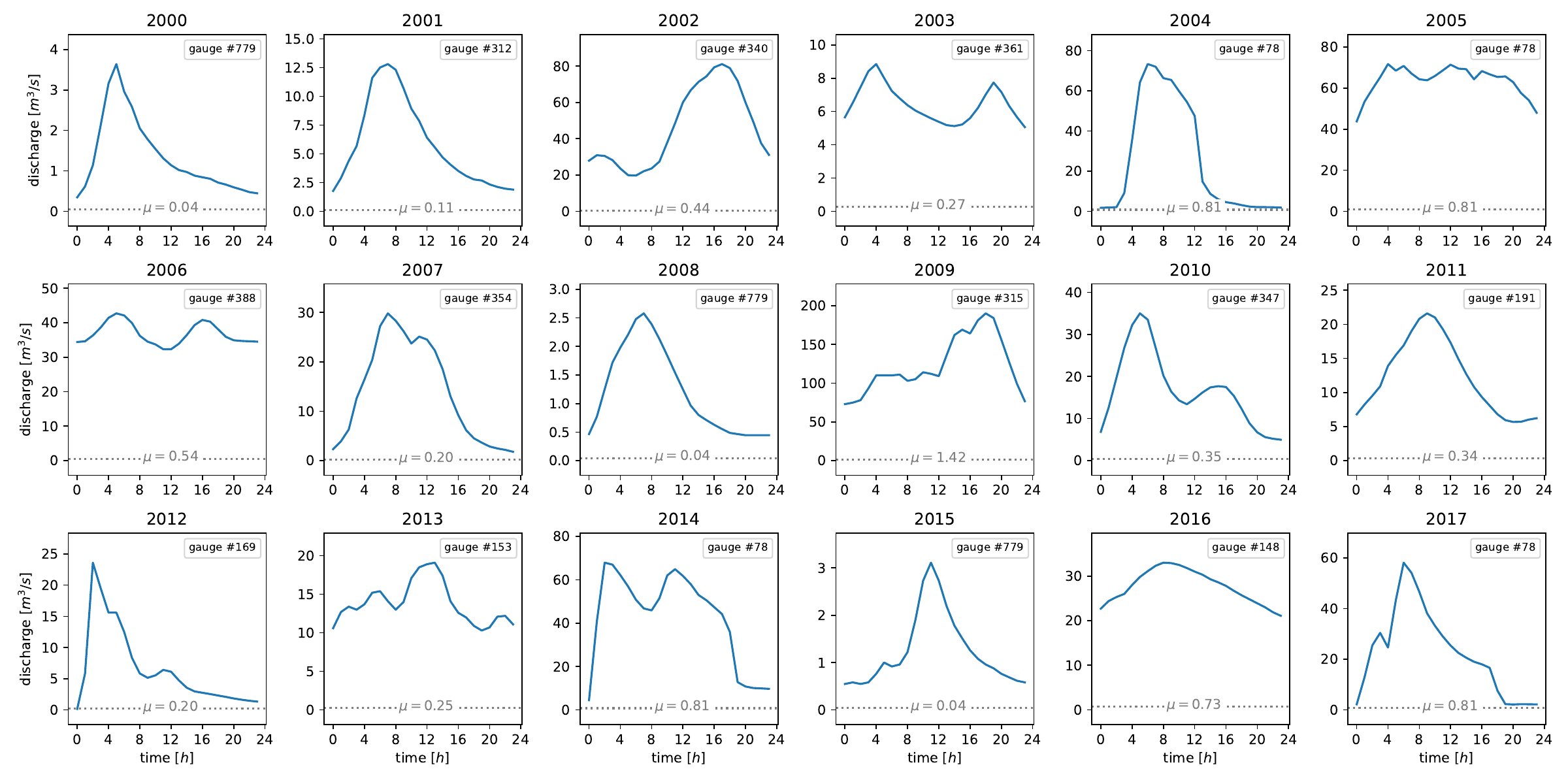}
    \caption{Year-wise maximisers of the relevancy score $\score$. Each maximiser's discharge window (blue) exhibits both high variability as well as an excessive overall discharge level, in relation to the mean discharge of its gauge (gray).}
\end{figure*}

\textbf{Relevancy Score.} The dataset contains many periods of almost no discharge activity. To guide the training process and focus on ``interesting" discharge windows, we seek to quantify the relevancy of each row $\bm{x}^{(g)} \in \R^W$ in the feature matrix $\featMat$. First, we \emph{unnormalise} to recover the original discharge values $\bm{x}^{(g)}_{\star} \coloneqq \sigma_g\bm{x}^{(g)} + \mu_g$. Then, let $\nabla\bm{x}^{(g)}_{\star}  \in \R^W$ denote the numerical derivative according to the second-order accurate central differences method, and ${\textstyle\int} \bm{x}^{(g)}_{\star}  \in \R$ the numerical integral according to the trapezoidal rule. We define the relevancy as\vspace*{-1mm}
\[
\score(\bm{x}^{(g)}) \coloneqq \operatorname{mean}\left(\frac{\nabla \bm{x}^{(g)}_{\star}}{\mu_g}\right)^2 \odot \frac{\textstyle\int \bm{x}^{(g)}_{\star} }{\mu_g}\in \R^W.
\]
This heuristic definition captures both the rate of change in a given discharge window and the overall discharge in relation to its mean, while weighting the former twice as strongly. The year-wise maximisers shown in \cref{fig:relevancy} suggest that this is a reasonable measure of relevancy. Note that it does not depend on the meteorological context variables.

\textbf{Optimisation Objective.} To measure the error between a model prediction $\predVec$ for input $\featMat$ and the target $\trueVec$, we weight the standard multi-dimensional regression square loss by the relevancy score:
\[
L(\predVec, \trueVec, \featMat) \coloneqq \tfrac{1}{n} \norm{\score(\featMat) \odot (\predVec - \trueVec)}_2^2.
\]
Training is then defined as optimising the expected loss over the empirical distribution of training samples in $\mathcal{D}_\text{train}$, regularised by the $\ell^2$-norm of the parameters:
\begin{equation*}
	\mathop{\min}_{\theta} \E_{(\featMat, \trueVec) \sim \mathcal{D}_\text{train}}[L(f_\theta(\featMat, \adjMatNorm), \trueVec, \featMat)] + \tfrac{\lambda}{2}\norm{\theta}_2^2.
\end{equation*}

\vspace*{-1mm}
\textbf{Testing Metric.} Recall that we perform training on normalised samples. For testing, we must calculate metrics on the unnormalised version of the predictions and targets:
\[
	\predVec_\star \coloneqq \stdVec \odot \predVec + \meanVec,\qquad
	\trueVec_\star \coloneqq \stdVec \odot \trueVec + \meanVec.
\]
The standard metric in hydrology for a single gauge is the \emph{Nash-Sutcliffe Efficiency} \citep[NSE;][]{nash_river_1970}. It compares the sum of squared errors of the model to the that of the constant mean-predictor and subtracts this value from one to obtain a percentage score in $[0,1]$. An NSE of zero means that the model's predictive capability is no better than that of the empirical mean, while an NSE of one indicates perfect model predictions. Since we are training the model with a weighted objective, we analogously weight\footnote{The unweighted version yields qualitatively similar results but has higher absolute values since it rewards performing well on trivial windows more than the weighted metric.} the evaluation metric with the relevancy score:\vspace*{-1mm}
\[
\mathbf{NSE} \coloneqq
1 - \frac{\sum_{i = 1}^{\abs{\mathcal{D}_\text{test}}} {\score(\featMat^{(t)}) \odot} (\predVec_\star^{(t_i)} - \trueVec_\star^{(t_i)})^2}{\sum_{i = 1}^{\abs{\mathcal{D}_\text{test}}} {\score(\featMat^{(t)}) \odot}  (\meanVec - \trueVec_\star^{(t_i)})^2}.
\]
We straightforwardly obtain a summary metric by averaging across gauges:
\(
\overline{\nse} \coloneqq \frac{1}{n} \sum_{g = 1}^{n} \nse_g.
\)
\section{Experiments}
\subsection{Experimental Setup} \label{sec:setup}

The code to reproduce our experiments is publicly available\footnote{\url{https://github.com/nkirschi/neural-flood-forecasting}}. \cref{tab:hyperparams} lists the relevant hyperparameters we use throughout all experiments unless stated otherwise.
On the data side, we set the window size to $W = 24$ and lead time to $L = 6$ hours, which are realistic choices. While, conceptually, larger window sizes would be preferred to provide a longer history to the predictor, they also imply a larger latent space dimensionality $d$ and thus restrict computational feasibility.

\begin{table}[h!]
 \vspace*{-8mm}
 	\caption{Default hyperparameters for our experiments.}
	\label{tab:hyperparams}
	\centering
	\begin{sc}
		\footnotesize
		\begin{tabular}{llr}
			\toprule
			& Hyperparameter & Value \\
			\midrule
			\multirow{3}{*}{\rotatebox{90}{data}} &
			window size ($W$) & \qty{24}{h} \\
			& lead time ($L$) & \qty{6}{h} \\
			& normalisation? & z-score \\
			\midrule
			\multirow{5}{*}{\rotatebox{90}{model}} &
			architecture & GCNII \\
			& network depth ($N$) & 19 \\
			& latent space dim ($d$) & 128 \\
			& edge direction & bidirected \\
			& adjacency type & binary \\
			\midrule
			\multirow{6}{*}{\rotatebox{90}{training}} &
			initialisation & kaiming \\
			& optimiser & adam \\
			& \# epochs    & 100 \\
			& batch size & 64 \\
			& learning rate & 10\textsuperscript{-4}\\
          & regularisation strength ($\lambda$) & 10\textsuperscript{-5} \\
			\bottomrule
		\end{tabular}
	\end{sc}
\end{table}

On the model side, we consider all three choices of layer definition described in \cref{sec:graphdef}, resulting in three model architectures ResGCN, GCNII, and ResGAT. We choose a depth of $N = 19$ layers to allow information propagation along the entire river graph, since the longest path in the preprocessed graph consists of 19 edges. The latent space dimensionality of $d = 128$ is chosen large enough to allow an injective feature embedding but small enough to avoid memory issues. The edge direction and adjacency type hyperparameters are subject to investigation in \cref{sec:riverstructureimpact}.

On the training side,  all neural network parameters are randomly initialised using the standard Kaiming initialisation scheme \citep{he_delving_2015} for architectures with ReLU activations. We then perform $100$ epochs of stochastic mini-batch gradient descent, which is enough for the process to converge. The descent algorithm is Adaptive Moments \citep[Adam;] []{kingma_adam_2015} with a learning rate of $10^{-4}$. To prevent overfitting, besides regularising with a strength of $\lambda = 10^{-5}$, we select the parameters from the epoch with minimal loss on a random holdout set containing $\frac{1}{5}$ of the training data.

\subsection{River Topology Comparison}
\label{sec:riverstructureimpact}

Our main experiment compares the impact of the six different gauge adjacency definitions detailed in \cref{sec:graphdef} on forecasting performance. In addition, we also consider three alternative edge orientations, which determine the direction of information flow in the GNN, as none of the options is a priori preferable. The \emph{downstream} orientation is given by the dataset, the \emph{upstream} orientation results from reversing all edges, and the \emph{bidirected} orientation from adding all reverse edges to the forward ones. We cross-validate all 18 topology combinations on the three train-test splits established in \cref{sec:preprocessing} using the summary NSE metric defined in \cref{sec:graphdef}, and report the results in \cref{tab:structure}.

\begin{table}[p]
\vspace*{-2mm}
\caption{Forecasting performance on different river network topologies, given as mean and standard deviation of $\overline{\nse}$ across folds. A wide 2-layer MLP baseline achieves a result of {$\qty{85.37}{\percent}\pm\qty{1.64}{\percent}$}.  Bold indicates the best value per column.  Note that results for the isolated adjacency type are not affected by the choice of edge orientation due to the absence of edges in this case.} 
\label{tab:structure}
\begin{subtable}[h]{\linewidth}
\vspace*{6mm}
    \caption{ResGCN}
    \newcommand{\error}[2]{\cell{\qty{#1}{}}{\qty{#2}{}}}
    \newcommand{\nash}[2]{\cell{\footnotesize\qty{#1}{\percent}}{\qty{#2}{\percent}}}
	\label{tab:topology_resgcn}
	\centering
	\begin{sc}
		\scriptsize
		\begin{tabular}{lrrr}
			\toprule
          \multirow{2}{*}{\vspace*{-1mm}adjacency type}& \multicolumn{3}{c}{edge orientation} \\ \cmidrule(lr){2-4}
			& downstream
			& upstream
			& bidirected \\
			\midrule
isolated & \boldmath\nash{85.07}{0.66} & \boldmath\nash{85.07}{0.66} & \boldmath\nash{85.07}{0.66}  \\ \midrule
binary &  \nash{82.03}{1.97} &  \nash{83.90}{1.26} &  \nash{82.73}{2.54}  \\ \midrule
stream length & \nash{81.64}{1.45} & \nash{81.98}{3.06} & \nash{83.09}{2.37}  \\ \cmidrule(l){1-4}
elevation difference & \nash{82.16}{1.85} & \nash{83.43}{0.16} & \nash{83.16}{1.76}  \\ \cmidrule(l){1-4}
average slope & \nash{81.93}{1.18} & \nash{80.68}{1.99} & \nash{81.59}{2.21}  \\ \midrule
learned & \nash{81.34}{1.61} & \nash{84.13}{0.81} &  \nash{83.50}{1.59}  \\
			\bottomrule
		\end{tabular}
	\end{sc}
 \end{subtable}

 \begin{subtable}[h]{\linewidth}
  \vspace*{6mm}
    \caption{GCNII}
    \newcommand{\error}[2]{\cell{\qty{#1}{}}{\qty{#2}{}}}
    \newcommand{\nash}[2]{\cell{\footnotesize\qty{#1}{\percent}}{\qty{#2}{\percent}}}
	\label{tab:topology_gcnii}
	\centering
	\begin{sc}
		\scriptsize
		\begin{tabular}{lrrr}
			\toprule
          \multirow{2}{*}{\vspace*{-1mm}adjacency type}& \multicolumn{3}{c}{edge orientation} \\ \cmidrule(lr){2-4}
			& downstream
			& upstream
			& bidirected \\
			\midrule
isolated & \nash{84.12}{1.88} & \nash{84.12}{1.88} & \nash{84.12}{1.88}  \\ \midrule
binary &  \nash{84.09}{1.11} &  \boldmath\nash{85.16}{1.74} &  \nash{84.81}{0.53}  \\ \midrule
stream length & \nash{84.29}{1.28} & \nash{85.09}{2.11} & \nash{83.90}{1.05}  \\ \cmidrule(l){1-4}
elevation difference & \nash{84.44}{0.81} & \nash{84.87}{1.78} & \nash{84.06}{0.68}  \\ \cmidrule(l){1-4}
average slope & \nash{83.93}{1.39} & \nash{84.47}{1.11} & \nash{84.68}{0.68}  \\ \midrule
learned & \boldmath\nash{84.91}{1.97} & \nash{85.00}{2.11} &  \boldmath\nash{85.56}{1.41}  \\
			\bottomrule
		\end{tabular}
	\end{sc}
 \end{subtable}

 \begin{subtable}[h]{\linewidth}
  \vspace*{6mm}
    \caption{ResGAT}
    \newcommand{\error}[2]{\cell{\qty{#1}{}}{\qty{#2}{}}}
    \newcommand{\nash}[2]{\cell{\footnotesize\qty{#1}{\percent}}{\qty{#2}{\percent}}}
	\label{tab:topology_resgat}
	\centering
	\begin{sc}
		\scriptsize
		\begin{tabular}{lrrr}
			\toprule
          \multirow{2}{*}{\vspace*{-1mm}adjacency type}& \multicolumn{3}{c}{edge orientation} \\ \cmidrule(lr){2-4}
			& downstream
			& upstream
			& bidirected \\
			\midrule
isolated & \nash{83.10}{0.88} & \nash{83.10}{0.88} & \nash{83.10}{0.88}  \\ \midrule
binary &  \nash{80.68}{4.78} &  \nash{82.59}{2.01} &  \nash{82.77}{0.47}  \\ \midrule
all of the below & \boldmath\nash{83.78}{1.71} & \boldmath\nash{83.33}{1.76} &  \boldmath\nash{82.73}{1.30}  \\\midrule
stream length & \nash{80.21}{4.85} & \nash{83.28}{1.72} & \nash{83.56}{1.57}  \\ \cmidrule(l){1-4}
elevation difference & \nash{80.58}{5.00} & \nash{82.88}{1.50} & \nash{82.87}{1.44}  \\ \cmidrule(l){1-4}
average slope & \nash{81.10}{4.67} & \nash{82.81}{0.90} & \nash{81.69}{0.39}  \\ 
			\bottomrule
		\end{tabular}
	\end{sc}
 \end{subtable}
\end{table}

\begin{table*}
	\caption{Pearson correlation between learned and physical edge weights.}

    \newcommand{\corr}[2]{\cell{\footnotesize\qty{#1}{}}{\qty{#2}{}}}
	\label{tab:correlation}
	\centering
	\begin{sc}
		\scriptsize
		\begin{tabular}{lrrrrrr}
			\toprule
			& \multicolumn{6}{c}{learned edge weights}\\ \cmidrule(lr){2-7}
         & \multicolumn{2}{c}{downstream} &  \multicolumn{2}{c}{upstream} &  \multicolumn{2}{c}{bidirected} \\\cmidrule(lr){2-3} \cmidrule(lr){4-5}\cmidrule(lr){6-7}
          physical edge weights & ResGCN & GCNII & ResGCN & GCNII & ResGCN & GCNII \\
			\midrule
			stream length & \corr{-0.375}{0.012} & \corr{-0.285}{0.014} & \corr{0.012}{0.056}  & \corr{0.027}{0.025} & \corr{0.139}{0.024} & \corr{-0.021}{0.048} \\\cmidrule(l){1-7}
			elevation difference & \corr{-0.148}{0.006} & \corr{-0.214}{0.013}  & \corr{-0.346}{0.025} & \corr{-0.325}{0.030}  & \corr{-0.182}{0.031} & \corr{-0.188}{0.051} \\\cmidrule(l){1-7}
			average slope & \corr{0.075}{0.007} & \corr{-0.034}{0.018}  & \corr{-0.325}{0.014} & \corr{-0.2955}{0.051} & \corr{-0.242}{0.017} & \corr{-0.158}{0.036} \\
			\bottomrule
		\end{tabular}
	\end{sc}
\end{table*}

Surprisingly, model performance shows almost no sensitivity to the choice of graph topology. Isolating the gauges does not harm performance beyond the standard deviation, and no combination outperforms a 19-layer MLP baseline. This indicates that the forecasting task for a gauge mainly benefits from the past discharge at that gauge but not from the discharge at neighbouring gauges. The river graph topology makes no difference. Even when the model is allowed to learn an optimal edge weight assignment, it does not manage to outperform the baseline.

\subsection{Learning the Weights}

The case of learned edge weights is of particular interest. They were initialised by drawing from the uniform distribution in $[0.9, 1.1]$ to arrange them neutrally around one while still introducing sufficient noise to break symmetry. Whenever learned weights get negative during training, we clip them to zero. The distribution of the learned weights (cp. \cref{tab:weightstats}) is still centred around one with minima close to zero and maxima below ten.

To see if the learned weights exhibit any similarities with the physical weights, we calculate Pearson correlation coefficients for all topology combinations. \cref{tab:correlation} shows that none of the physical weight assignments correlate much with the learned weights. In multiple instances, the sign even flips when using a different model architecture. For instance, the largest positive correlation occurs with stream length for ResGCN, but in this same case GCNII achieves a negative correlation of the same magnitude. Hence, we conclude that none of the physical edge weights from the datasets are optimal context information for the predictor.

\subsection{Small-scale Subnetworks}

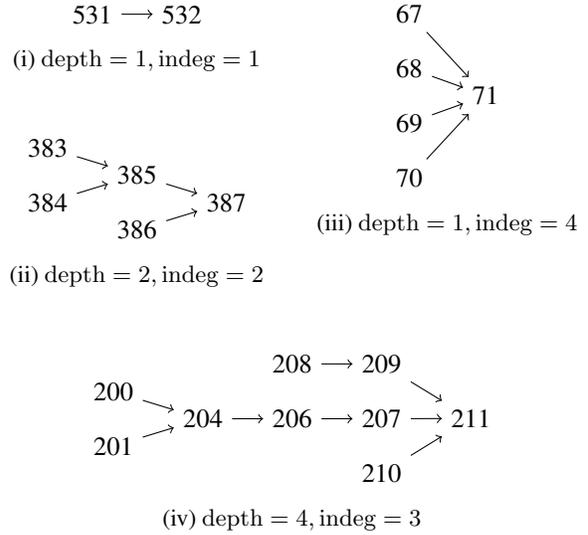
\begin{figure}[p]
\begin{minipage}[c]{0.5\linewidth}
\begin{subfigure}[b]{\linewidth}
\label{fig:subgraph1}
\centering
\begin{forest}
  my box/.style={top color=blue!20, draw, rounded corners, bottom color=blue!30},
  for tree={
    grow=west,
    edge+={<-},
  },
  [532
    [531]
  ]
\end{forest}
\caption{$\mathrm{depth} = 1, \mathrm{indeg} = 1$}
\end{subfigure}
\begin{subfigure}[b]{\linewidth}
\label{fig:subgraph2}
\centering
\vspace*{7mm}
\begin{forest}
  my box/.style={top color=blue!20, draw, rounded corners, bottom color=blue!30},
  for tree={
    grow=west,
    edge+={<-},
  },
  [387
    [385
        [383]
        [384]
    ]
    [386]
  ]
\end{forest}
\caption{$\mathrm{depth} = 2, \mathrm{indeg} = 2$}
\end{subfigure}
\end{minipage}\begin{minipage}[c]{0.5\linewidth}
\begin{subfigure}[b]{\linewidth}
\label{fig:subgraph3}
\centering
\begin{forest}
  my box/.style={top color=blue!20, draw, rounded corners, bottom color=blue!30},
  for tree={
    grow=west,
    edge+={<-},
  },
  [71
    [67]
    [68]
    [69]
    [70]
  ]
\end{forest}
\caption{$\mathrm{depth} = 1, \mathrm{indeg} = 4$}
\vspace*{7mm}
\end{subfigure}
\end{minipage}
\begin{subfigure}[b]{\linewidth}
\label{fig:subgraph4}
\centering
\vspace*{7mm}
\begin{forest}
  my box/.style={top color=blue!20, draw, rounded corners, bottom color=blue!30},
  for tree={
    grow=west,
    edge+={<-},
  },
  [211
    [209
        [208, tier=t2]
    ]
    [207
        [206
            [204
                [200]
                [201]
            205]
        ]
    ]
    [210, tier=t1]
  ]
\end{forest}
\caption{$\mathrm{depth} = 4, \mathrm{indeg} = 3$}
\end{subfigure}
\caption{Four subgraphs of the river network with different depth and sink in-degree. The node labels refer to the original gauge IDs from the dataset.}
\label{fig:subnetworks}
\end{figure}

To exclude the possibility that the considerable depth is causing the GCN to not outperform the baseline MLP due to more general issues with training very deep networks, we repeat the topology comparison from \cref{sec:riverstructureimpact} on four small subgraphs of the river network illustrated in \cref{fig:subnetworks}. Since the graph rewiring done by \cref{algo:softRemoval} can have a strong effect when considering only a handful nodes, we skipped it in the preprocessing for this experiment and chose only subgraphs with full data coverage to begin with. Furthermore, to allow for sufficient model capacity, we increase the latent space dimensionality to 512 for this experiment.

The results are consistent with those on the full dataset and hence outsourced into the appendix tables \ref{tab:subgraph1} to \ref{tab:subgraph4}. Note that the orders of magnitude of $\overline{\nse}$ differ as the difficulty of the prediction task naturally changes with the underlying graph. The small-scale experiment confirms the observation that topology context does not benefit prediction.

\renewcommand\thesubfigure{\roman{subfigure}}
\renewcommand\thesubfigure{\alpha{subfigure}}

\subsection{Worst Gauge Investigation} \label{sec:outlier}

\begin{figure}[p]
	\centering
	\includegraphics[width=0.5\textwidth]{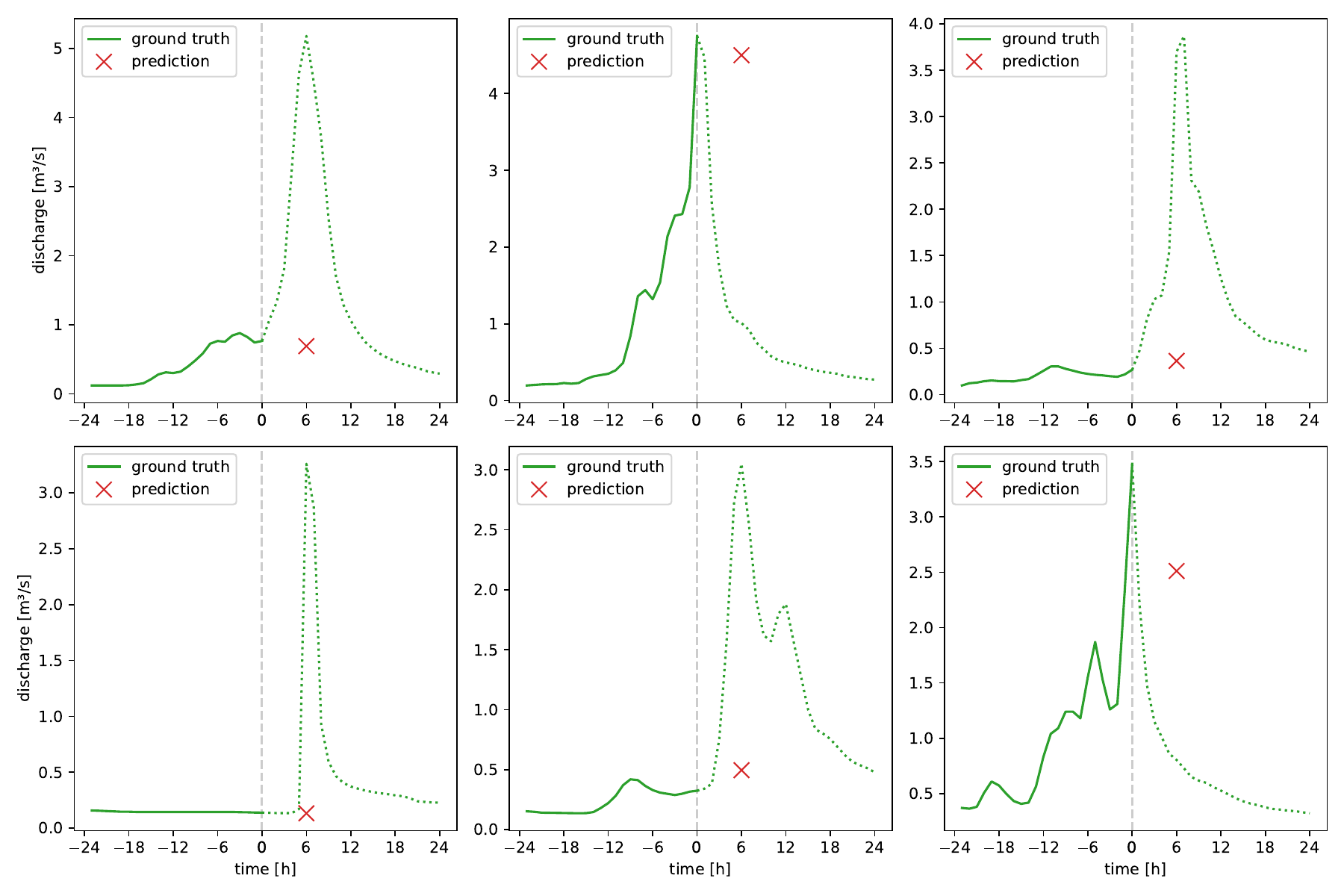}
	\caption{Worst predictions of the bidirected-learned GCNII (third fold) on its overall worst gauge \#169. Negative time indicates past, and positive time indicates future discharge.}
	\label{fig:recurrent_forecasting}
\end{figure}

The performance on gauge \#169 of all trained models is considerably below the mean. For instance, the best overall performing model, bidirected-learned GCNII (third fold), achieves its worst $\overline{\nse}$ on this outlier gauge. To better understand the scenarios that are challenging for the model, we determine the top disjoint time horizons of 48 hours (24 hours for past and future) in terms of deviation of model prediction from the ground truth. The resulting plots in \cref{fig:recurrent_forecasting} reveal that the outlier gauge is characterised by sudden and narrow spikes, which are inherently hard to forecast for any predictor. The gauge might be located behind a floodgate. As a result, the forecasting performance is mediocre, with the forecast often missing spikes.

\section{Conclusion}

In this work, we explored the applicability of GNNs to holistic flood forecasting in a river network graph. Based on the LamaH-CE dataset, we framed a supervised node regression task for predicting future discharge at all gauging stations in the graph given past observations. By modifying the adjacency matrix, we compared the impact of different adjacency definitions on the prediction performance. Our results reveal that the impact of river topology is negligible. The GNN performs equally well even when all edges are removed from the graph, which makes it act like an MLP. It does not benefit from weighted edges that resemble physical relationships between gauges. When the model is allowed to jointly learn the edge weights along with the other parameters, they correlate with neither constant weights nor any of the physical weightings given by the dataset. A small-scale subnetwork study shows that the results are not caused by issues with training deep models but prove consistent throughout scales. Investigations on a challenging outlier gauge indicate that the GNNs struggle to predict sudden, narrow discharge spikes.

On a high level, future work is encouraged to investigate under which conditions including graph topology in neural predictors actually helps, which is not clear a priori. While the key could lie in employing more specialised model architectures such as DAGNN \citep{thost_directed_2021} for the dataset at hand, there might be more fundamental limitations to the use of GNNs for large-scale regression problems. 
Moreover, for the application of flood forecasting, enhancing the dataset with inter-gauge propagation time metadata and reliable flood annotations may allow the predictor to leverage the relational context more effectively. Otherwise, our results suggest that focusing on accurate spike prediction might be more promising than incorporating river network topology information.

Finally, there is a broader issue: we used a river network dataset from central Europe as discharge measurements are readily available there for long time periods. However, the regions most affected by floods happen to be typically located in low-income countries where data is scarce. More gauges need to be installed in those high-risk regions, and large-scale datasets collected to enable more relevant studies and save lives.

\section*{Impact Statement}
Flood forecasting is a crucial technology to mitigate humanitarian crises in the age of climate change. With floods being the most frequent type of natural disaster, even minor methodological improvements are bound to greatly impact disaster prevention and mitigation. While the results in our work suggest that incorporating river network topology into the forecasting process might not be one such improvement, the momentousness of the topic demands that future research continues to explore the idea as well as conditions under which it potentially can be an improvement.

\section*{Acknowledgements}
 This work was supported by a fellowship of the German Academic Exchange Service (DAAD) and funding from the Centre for Doctoral Training in Industrially Focused Mathematical Modelling.
 We thank Professor Terry Lyons and the Mathematical Institute, University of Oxford for providing computing resources and Dr Frederik Kratzert from Google Research for helpful discussions.

 \newpage

\bibliography{literature}
\bibliographystyle{icml2024}

\newpage
\appendix
\onecolumn
\section{Appendix}

\subsection{Preprocessing Algorithms}

 \begin{minipage}[t]{.43\linewidth}
\begin{algorithm}[H]
	\KwIn{DAG $\graph = (\vertices, \edges)$, start node $v_0 \in \vertices$}
	\KwOut{All direct and indirect predecessors of $v_0$ in $\graph$}
	\SetKwBlock{Begin}{}{}
	\vspace{5mm}
	\nonl\Begin($\mathtt{inverseDFS} {(} \graph, v_0 {)}$)
	{
		$\predecessors \gets \{v \in \vertices \mid (v, v_0) \in \edges\}$
		
		\If{$\predecessors = \emptyset$}{
			\Return $\{v_0\}$
		}
		\Else{
			\Return $\{v_0\} \cup \displaystyle\bigcup_{v \in \predecessors} \mathtt{invDFS}(v)$
		}
	}
	\caption{Inverse depth-first search}
	\label{algo:invserseDFS}
\end{algorithm}
\end{minipage}\hfill\begin{minipage}[t]{.53\linewidth}
\begin{algorithm}[H]
	\newcommand{\vRIP}{v_{\text{\scshape rip}}}
	\KwIn{antitransitive weighted DAG $\graph = (\vertices, \edges, w)$, moribund node $\vRIP \in \vertices$}
	\KwOut{$\graph$ without $\vRIP$ where its predecessors and successors are rewired}
	\SetKwBlock{Begin}{}{}
	\vspace{5mm}
	\nonl\Begin($\mathtt{rewireRemove} {(} \graph, \vRIP {)}$)
	{
		$\predecessors \gets \{v \in \vertices \mid (v, \vRIP) \in \edges\}$
		
		$\successors \gets \{v \in \vertices \mid (\vRIP, v) \in \edges\}$
		
		$\vertices \gets \vertices \setminus \{\vRIP\}$
		
		$\edges \gets \edges \setminus (\predecessors \times \{\vRIP\}) \setminus (\{\vRIP\} \times \successors) \cup (\predecessors \times \successors)$

    \For{$(v_\mathrm{in}, v_\mathrm{out}) \in \predecessors \times \successors$}{
     $w(v_\mathrm{in}, v_\mathrm{out}) \gets w(v_\mathrm{in}, \vRIP) + w(\vRIP, v_\mathrm{out})$
     }
	}
	\caption{Rewire-removal of a node}
	\label{algo:softRemoval}
\end{algorithm}
\end{minipage}

\vspace*{1cm}

\subsection{Learned Edge Weights Statistics}

\begin{table*}[h]
\renewcommand\thetable{A.3}
	\caption{Key statistics of the learned edge weights, accumulated across folds.}
	\label{tab:weightstats}
    \newcommand{\corr}[2]{\cell{\qty{#1}{}}{\qty{#2}{}}}

	\centering
	\begin{sc}
		\footnotesize
		\begin{tabular}{lrrrrrr}
			\toprule
         & \multicolumn{2}{c}{downstream} &  \multicolumn{2}{c}{upstream} &  \multicolumn{2}{c}{bidirectional} \\\cmidrule(lr){2-3} \cmidrule(lr){4-5}\cmidrule(lr){6-7}
          statistic & ResGCN & GCNII & ResGCN & GCNII & ResGCN & GCNII \\
			\midrule
			mean & \corr{0.462}{0.082} & \corr{0.263}{0.072} & \corr{0.670}{0.039}  & \corr{0.627}{0.056} & \corr{0.789}{0.044} & \corr{0.618}{0.062} \\\cmidrule(l){1-7}
			std & \corr{0.322}{0.013} & \corr{0.281}{0.038}  & \corr{0.375}{0.004} & \corr{0.369}{0.013}  & \corr{0.329}{0.008} & \corr{0.361}{0.005} \\\midrule
			min & \corr{0.000}{0.000} & \corr{0.000}{0.000}  & \corr{0.000}{0.000} & \corr{0.000}{0.000} & \corr{0.061}{0.045} & \corr{0.000}{0.000} \\\cmidrule(l){1-7}
			25\% & \corr{0.191}{0.086} & \corr{0.033}{0.033}  & \corr{0.382}{0.049} & \corr{0.345}{0.066} & \corr{0.556}{0.039} & \corr{0.342}{0.079} \\\cmidrule(l){1-7}
			median & \corr{0.416}{0.084} & \corr{0.158}{0.091}  & \corr{0.708}{0.064} & \corr{0.628}{0.075} & \corr{0.802}{0.049} & \corr{0.592}{0.066} \\\cmidrule(l){1-7}
			75 \% & \corr{0.689}{0.102} & \corr{0.434}{0.092}  & \corr{0.959}{0.036} & \corr{0.894}{0.078} & \corr{1.032}{0.040} & \corr{0.901}{0.075} \\\cmidrule(l){1-7}
			max \% & \corr{1.313}{0.089} & \corr{1.376}{0.094}  & \corr{1.471}{0.052} & \corr{1.609}{0.102} & \corr{1.565}{0.037} & \corr{1.668}{0.083} \\
			\bottomrule
		\end{tabular}
	\end{sc}
\end{table*}


\newpage
\twocolumn
\subsection{Subnetwork Results}
\label{sec:gadgets}
\vspace*{0.7mm}

\begin{table}[h!] 
\renewcommand\thetable{A.4}
\caption{Cross-validation $\overline{\nse}$ on subgraph (i) in \cref{fig:subnetworks}.}
\label{tab:subgraph1}
\begin{subtable}[h]{\linewidth}
\vspace*{6mm}
    \caption{ResGCN}
    \newcommand{\error}[2]{\cell{\qty{#1}{}}{\qty{#2}{}}}
    \newcommand{\nash}[2]{\cell{\footnotesize\qty{#1}{\percent}}{\qty{#2}{\percent}}}
	
	\centering
	\begin{sc}
		\scriptsize
		\begin{tabular}{lrrr}
			\toprule
          \multirow{2}{*}{\vspace*{-1mm}adjacency type}& \multicolumn{3}{c}{edge orientation} \\ \cmidrule(lr){2-4}
			& downstream
			& upstream
			& bidirected \\
			\midrule
isolated & \nash{75.07}{7.25} & \nash{75.07}{ 7.25} & \nash{75.07}{7.25}  \\ \midrule
binary &  \nash{82.77}{1.26} &  \nash{81.20}{3.68} &  \nash{77.61}{4.43}  \\ \midrule
stream length & \nash{82.77}{1.26} & \nash{81.20}{3.68} & \nash{77.61}{4.43}  \\ \cmidrule(l){1-4}
elevation difference & \nash{81.76}{3.11} & \nash{81.42}{3.74} & \nash{77.31}{4.18}  \\ \cmidrule(l){1-4}
average slope & \nash{81.76}{3.11} & \nash{81.42}{3.74} & \nash{77.31}{4.18}  \\ \midrule
learned & \nash{81.82}{5.63} & \nash{81.61}{2.63} &  \nash{75.47}{7.02}  \\
			\bottomrule
		\end{tabular}
	\end{sc}
 \end{subtable}

 \begin{subtable}[h!]{\linewidth}
  \vspace*{6mm}
    \caption{GCNII}
    \newcommand{\error}[2]{\cell{\qty{#1}{}}{\qty{#2}{}}}
    \newcommand{\nash}[2]{\cell{\footnotesize\qty{#1}{\percent}}{\qty{#2}{\percent}}}
	
	\centering
	\begin{sc}
		\scriptsize
			\begin{tabular}{lrrr}
			\toprule
          \multirow{2}{*}{\vspace*{-1mm}adjacency type}& \multicolumn{3}{c}{edge orientation} \\ \cmidrule(lr){2-4}
			& downstream
			& upstream
			& bidirected \\
			\midrule
isolated & \nash{80.44}{3.16} & \nash{80.44}{3.16} & \nash{80.44}{3.16}  \\ \midrule
binary &  \nash{80.52}{4.53} &  \nash{74.57}{2.27} &  \nash{81.16}{2.97}  \\ \midrule
stream length & \nash{80.52}{4.53} & \nash{74.57}{2.27} & \nash{81.16}{2.97}  \\ \cmidrule(l){1-4}
elevation difference & \nash{78.22}{4.74} & \nash{74.42}{4.53} & \nash{80.30}{4.11}  \\ \cmidrule(l){1-4}
average slope & \nash{78.22}{4.74} & \nash{74.42}{4.53} & \nash{80.30}{4.11}  \\ \midrule
learned & \nash{80.82}{4.05} & \nash{75.67}{4.29} &  \nash{79.99}{7.83}  \\
			\bottomrule
		\end{tabular}
	\end{sc}
 \end{subtable}

 \begin{subtable}[h]{\linewidth}
  \vspace*{6mm}
    \caption{ResGAT}
    \newcommand{\error}[2]{\cell{\qty{#1}{}}{\qty{#2}{}}}
    \newcommand{\nash}[2]{\cell{\footnotesize\qty{#1}{\percent}}{\qty{#2}{\percent}}}
	
	\centering
	\begin{sc}
		\scriptsize
				\begin{tabular}{lrrr}
			\toprule
          \multirow{2}{*}{\vspace*{-1mm}adjacency type}& \multicolumn{3}{c}{edge orientation} \\ \cmidrule(lr){2-4}
			& downstream
			& upstream
			& bidirected \\
			\midrule
isolated & \nash{76.81}{9.40} & \nash{76.81}{9.40} & \nash{76.81}{9.40}  \\ \midrule
binary &  \nash{82.28}{6.65} &  \nash{83.85}{3.56} &  \nash{77.47}{6.20}  \\ \midrule
all of the below & \nash{75.60}{7.43} & \nash{76.56}{10.01} &  \nash{75.88}{9.85}  \\\midrule
stream length & \nash{82.28}{6.65} & \nash{83.85}{3.56} & \nash{77.47}{6.20}  \\ \cmidrule(l){1-4}
elevation difference & \nash{82.28}{6.65} & \nash{83.85}{3.56} & \nash{77.47}{6.20}  \\ \cmidrule(l){1-4}
average slope & \nash{82.28}{6.65} & \nash{83.85}{3.56} & \nash{77.47}{6.20}  \\ 
			\bottomrule
		\end{tabular}
	\end{sc}
 \end{subtable}

\end{table}

\begin{table}[h!] 
\renewcommand\thetable{A.5}
\caption{Cross-validation $\overline{\nse}$ on subgraph (ii) in \cref{fig:subnetworks}.}
\label{tab:subgraph2}
\begin{subtable}[h]{\linewidth}
\vspace*{6mm}
    \caption{ResGCN}
    \newcommand{\error}[2]{\cell{\qty{#1}{}}{\qty{#2}{}}}
    \newcommand{\nash}[2]{\cell{\footnotesize\qty{#1}{\percent}}{\qty{#2}{\percent}}}
	
	\centering
	\begin{sc}
		\scriptsize
		\begin{tabular}{lrrr}
			\toprule
          \multirow{2}{*}{\vspace*{-1mm}adjacency type}& \multicolumn{3}{c}{edge orientation} \\ \cmidrule(lr){2-4}
			& downstream
			& upstream
			& bidirected \\
			\midrule
isolated & \nash{93.20}{0.23} & \nash{93.20}{0.23} & \nash{93.20}{0.23}  \\ \midrule
binary &  \nash{92.61}{5.50} &  \nash{95.36}{1.45} &  \nash{95.40}{1.80}  \\ \midrule
stream length & \nash{92.71}{5.55} & \nash{94.82}{1.80} & \nash{94.30}{0.74}  \\ \cmidrule(l){1-4}
elevation difference & \nash{92.79}{5.56} & \nash{94.92}{1.49} & \nash{95.03}{1.81}  \\ \cmidrule(l){1-4}
average slope & \nash{92.68}{5.56} & \nash{95.28}{0.94} & \nash{95.25}{0.58}  \\ \midrule
learned & \nash{92.69}{5.79} & \nash{96.17}{0.25} &  \nash{96.12}{0.92}  \\
			\bottomrule
		\end{tabular}
	\end{sc}
 \end{subtable}

 \begin{subtable}[h]{\linewidth}
  \vspace*{6mm}
    \caption{GCNII}
    \newcommand{\error}[2]{\cell{\qty{#1}{}}{\qty{#2}{}}}
    \newcommand{\nash}[2]{\cell{\footnotesize\qty{#1}{\percent}}{\qty{#2}{\percent}}}
	
	\centering
	\begin{sc}
		\scriptsize
			\begin{tabular}{lrrr}
			\toprule
          \multirow{2}{*}{\vspace*{-1mm}adjacency type}& \multicolumn{3}{c}{edge orientation} \\ \cmidrule(lr){2-4}
			& downstream
			& upstream
			& bidirected \\
			\midrule
isolated & \nash{91.95}{1.89} & \nash{91.95}{1.89} & \nash{91.95}{1.89}  \\ \midrule
binary &  \nash{95.61}{1.18} &  \nash{96.01}{0.70} &  \nash{93.65}{3.67}  \\ \midrule
stream length & \nash{95.50}{1.64} & \nash{95.80}{0.88} & \nash{93.05}{2.57}  \\ \cmidrule(l){1-4}
elevation difference & \nash{95.06}{1.03} & \nash{96.24}{0.75} & \nash{92.48}{3.43}  \\ \cmidrule(l){1-4}
average slope & \nash{95.58}{1.66} & \nash{96.05}{0.55} & \nash{95.23}{0.52}  \\ \midrule
learned & \nash{94.78}{1.27} & \nash{94.84}{1.54} &  \nash{94.23}{0.33}  \\
			\bottomrule
		\end{tabular}
	\end{sc}
 \end{subtable}

 \begin{subtable}[h]{\linewidth}
  \vspace*{6mm}
    \caption{ResGAT}
    \newcommand{\error}[2]{\cell{\qty{#1}{}}{\qty{#2}{}}}
    \newcommand{\nash}[2]{\cell{\footnotesize\qty{#1}{\percent}}{\qty{#2}{\percent}}}
	
	\centering
	\begin{sc}
		\scriptsize
				\begin{tabular}{lrrr}
			\toprule
          \multirow{2}{*}{\vspace*{-1mm}adjacency type}& \multicolumn{3}{c}{edge orientation} \\ \cmidrule(lr){2-4}
			& downstream
			& upstream
			& bidirected \\
			\midrule
isolated & \nash{92.36}{0.96} & \nash{92.36}{0.96} & \nash{92.36}{0.96}  \\ \midrule
binary &  \nash{93.40}{4.02} &  \nash{95.08}{1.33} &  \nash{94.43}{2.11}  \\ \midrule
all of the below & \nash{94.78}{0.81} & \nash{93.08}{2.80} &  \nash{94.47}{3.76}  \\\midrule
stream length & \nash{93.40}{4.02} & \nash{95.08}{1.33} & \nash{94.43}{2.11}  \\ \cmidrule(l){1-4}
elevation difference & \nash{93.40}{4.02} & \nash{95.08}{1.33} & \nash{94.43}{2.11}  \\ \cmidrule(l){1-4}
average slope & \nash{93.40}{4.02} & \nash{95.08}{1.33} & \nash{94.43}{2.11}  \\ 
			\bottomrule
		\end{tabular}
	\end{sc}
 \end{subtable}

\end{table}

\begin{table}[h!] 
\renewcommand\thetable{A.6}
\caption{Cross-validation $\overline{\nse}$ on subgraph (iii) in \cref{fig:subnetworks}.}
\label{tab:subgraph3}
\begin{subtable}[h]{\linewidth}
\vspace*{6mm}
    \caption{ResGCN}
    \newcommand{\error}[2]{\cell{\qty{#1}{}}{\qty{#2}{}}}
    \newcommand{\nash}[2]{\cell{\footnotesize\qty{#1}{\percent}}{\qty{#2}{\percent}}}
	
	\centering
	\begin{sc}
		\scriptsize
		\begin{tabular}{lrrr}
			\toprule
          \multirow{2}{*}{\vspace*{-1mm}adjacency type}& \multicolumn{3}{c}{edge orientation} \\ \cmidrule(lr){2-4}
			& downstream
			& upstream
			& bidirected \\
			\midrule
isolated & \nash{78.12}{3.80} & \nash{78.12}{3.80} & \nash{78.12}{3.80}  \\ \midrule
binary &  \nash{79.23}{3.07} &  \nash{81.77}{1.30} &  \nash{78.10}{1.66}  \\ \midrule
stream length & \nash{79.23}{3.07} & \nash{81.66}{1.18} & \nash{76.05}{2.27}  \\ \cmidrule(l){1-4}
elevation difference & \nash{79.23}{3.07} & \nash{81.81}{1.34} & \nash{76.66}{2.48}  \\ \cmidrule(l){1-4}
average slope & \nash{79.23}{3.07} & \nash{81.62}{1.05} & \nash{77.51}{2.02}  \\ \midrule
learned & \nash{77.72}{4.22} & \nash{82.39}{1.35} &  \nash{77.89}{3.52}  \\
			\bottomrule
		\end{tabular}
	\end{sc}
 \end{subtable}

 \begin{subtable}[h]{\linewidth}
  \vspace*{6mm}
    \caption{GCNII}
    \newcommand{\error}[2]{\cell{\qty{#1}{}}{\qty{#2}{}}}
    \newcommand{\nash}[2]{\cell{\footnotesize\qty{#1}{\percent}}{\qty{#2}{\percent}}}
	
	\centering
	\begin{sc}
		\scriptsize
			\begin{tabular}{lrrr}
			\toprule
          \multirow{2}{*}{\vspace*{-1mm}adjacency type}& \multicolumn{3}{c}{edge orientation} \\ \cmidrule(lr){2-4}
			& downstream
			& upstream
			& bidirected \\
			\midrule
isolated & \nash{79.58}{1.78} & \nash{79.58}{1.78} & \nash{79.58}{1.78}  \\ \midrule
binary &  \nash{78.64}{2.26} &  \nash{77.31}{0.86} &  \nash{77.60}{2.24}  \\ \midrule
stream length & \nash{77.52}{4.23} & \nash{75.38}{1.89} & \nash{78.92}{1.52}  \\ \cmidrule(l){1-4}
elevation difference & \nash{77.52}{4.23} & \nash{76.93}{2.07} & \nash{76.93}{0.90}  \\ \cmidrule(l){1-4}
average slope & \nash{78.64}{2.26} & \nash{76.13}{2.35} & \nash{79.46}{1.23}  \\ \midrule
learned & \nash{78.81}{2.84} & \nash{76.84}{1.38} &  \nash{79.04}{2.30}  \\
			\bottomrule
		\end{tabular}
	\end{sc}
 \end{subtable}

 \begin{subtable}[h]{\linewidth}
  \vspace*{6mm}
    \caption{ResGAT}
    \newcommand{\error}[2]{\cell{\qty{#1}{}}{\qty{#2}{}}}
    \newcommand{\nash}[2]{\cell{\footnotesize\qty{#1}{\percent}}{\qty{#2}{\percent}}}
	
	\centering
	\begin{sc}
		\scriptsize
				\begin{tabular}{lrrr}
			\toprule
          \multirow{2}{*}{\vspace*{-1mm}adjacency type}& \multicolumn{3}{c}{edge orientation} \\ \cmidrule(lr){2-4}
			& downstream
			& upstream
			& bidirected \\
			\midrule
isolated & \nash{81.21}{2.70} & \nash{81.21}{2.70} & \nash{81.21}{2.70}  \\ \midrule
binary &  \nash{76.35}{4.39} &  \nash{76.90}{1.48} &  \nash{77.30}{3.02}  \\ \midrule
all of the below & \nash{80.69}{2.65} & \nash{73.19}{11.11} &  \nash{75.19}{2.83}  \\\midrule
stream length & \nash{76.35}{4.39} & \nash{76.90}{1.48} & \nash{77.30}{3.02}  \\ \cmidrule(l){1-4}
elevation difference & \nash{76.35}{4.39} & \nash{76.90}{1.48} & \nash{77.30}{3.02}  \\ \cmidrule(l){1-4}
average slope & \nash{76.35}{4.39} & \nash{76.90}{1.48} & \nash{77.30}{3.02}  \\ 
			\bottomrule
		\end{tabular}
	\end{sc}
 \end{subtable}
\end{table}

\begin{table}[h!] 
\renewcommand\thetable{A.7}
\caption{Cross-validation $\overline{\nse}$ on subgraph (iv) in \cref{fig:subnetworks}.}
\label{tab:subgraph4}
\begin{subtable}[h]{\linewidth}
\vspace*{6mm}
    \caption{ResGCN}
    \newcommand{\error}[2]{\cell{\qty{#1}{}}{\qty{#2}{}}}
    \newcommand{\nash}[2]{\cell{\footnotesize\qty{#1}{\percent}}{\qty{#2}{\percent}}}
	
	\centering
	\begin{sc}
		\scriptsize
		\begin{tabular}{lrrr}
			\toprule
          \multirow{2}{*}{\vspace*{-1mm}adjacency type}& \multicolumn{3}{c}{edge orientation} \\ \cmidrule(lr){2-4}
			& downstream
			& upstream
			& bidirected \\
			\midrule
isolated & \nash{96.00}{0.43} & \nash{96.00}{0.43} & \nash{96.00}{0.43}  \\ \midrule
binary &  \nash{94.99}{0.25} &  \nash{96.20}{0.94} &  \nash{95.58}{0.59}  \\ \midrule
stream length & \nash{95.17}{0.27} & \nash{96.10}{1.12} & \nash{95.68}{0.54}  \\ \cmidrule(l){1-4}
elevation difference & \nash{95.01}{0.24} & \nash{96.18}{0.88} & \nash{95.71}{0.87}  \\ \cmidrule(l){1-4}
average slope & \nash{95.14}{0.24} & \nash{96.09}{1.02} & \nash{95.20}{0.85}  \\ \midrule
learned & \nash{95.11}{0.15} & \nash{96.34}{0.87} &  \nash{95.72}{0.70}  \\
			\bottomrule
		\end{tabular}
	\end{sc}
 \end{subtable}

 \begin{subtable}[h]{\linewidth}
  \vspace*{6mm}
    \caption{GCNII}
    \newcommand{\error}[2]{\cell{\qty{#1}{}}{\qty{#2}{}}}
    \newcommand{\nash}[2]{\cell{\footnotesize\qty{#1}{\percent}}{\qty{#2}{\percent}}}
	
	\centering
	\begin{sc}
		\scriptsize
			\begin{tabular}{lrrr}
			\toprule
          \multirow{2}{*}{\vspace*{-1mm}adjacency type}& \multicolumn{3}{c}{edge orientation} \\ \cmidrule(lr){2-4}
			& downstream
			& upstream
			& bidirected \\
			\midrule
isolated & \nash{95.90}{0.09} & \nash{95.90}{0.09} & \nash{95.90}{0.09}  \\ \midrule
binary &  \nash{96.05}{0.12} &  \nash{96.18}{0.45} &  \nash{96.02}{0.52}  \\ \midrule
stream length & \nash{95.93}{0.13} & \nash{96.26}{0.46} & \nash{96.12}{0.60}  \\ \cmidrule(l){1-4}
elevation difference & \nash{96.05}{0.08} & \nash{96.09}{0.48} & \nash{96.14}{0.57}  \\ \cmidrule(l){1-4}
average slope & \nash{96.02}{0.06} & \nash{96.15}{0.52} & \nash{95.92}{0.63}  \\ \midrule
learned & \nash{95.86}{0.25} & \nash{96.22}{0.63} &  \nash{96.21}{0.37}  \\
			\bottomrule
		\end{tabular}
	\end{sc}
 \end{subtable}

 \begin{subtable}[h]{\linewidth}
  \vspace*{6mm}
    \caption{ResGAT}
    \newcommand{\error}[2]{\cell{\qty{#1}{}}{\qty{#2}{}}}
    \newcommand{\nash}[2]{\cell{\footnotesize\qty{#1}{\percent}}{\qty{#2}{\percent}}}
	
	\centering
	\begin{sc}
		\scriptsize
				\begin{tabular}{lrrr}
			\toprule
          \multirow{2}{*}{\vspace*{-1mm}adjacency type}& \multicolumn{3}{c}{edge orientation} \\ \cmidrule(lr){2-4}
			& downstream
			& upstream
			& bidirected \\
			\midrule
isolated & \nash{96.01}{0.34} & \nash{96.01}{0.34} & \nash{96.01}{0.34}  \\ \midrule
binary &  \nash{95.19}{0.09} &  \nash{96.50}{0.06} &  \nash{95.90}{0.39}  \\ \midrule
all of the below & \nash{96.17}{0.26} & \nash{96.04}{0.26} &  \nash{95.89}{0.04}  \\\midrule
stream length & \nash{95.19}{0.09} & \nash{96.50}{0.06} & \nash{95.90}{0.39}  \\ \cmidrule(l){1-4}
elevation difference & \nash{95.19}{0.09} & \nash{96.50}{0.06} & \nash{95.90}{0.39}  \\ \cmidrule(l){1-4}
average slope & \nash{95.19}{0.09} & \nash{96.50}{0.06} & \nash{95.90}{0.39}  \\ 
			\bottomrule
		\end{tabular}
	\end{sc}
 \end{subtable}
\end{table}

\clearpage
\onecolumn

\subsection{Effect of Window Size and Lead Time}

\begin{table}[h!] 
\centering
\renewcommand\thetable{A.8}

\parbox{11.2cm}{\caption{Cross-validation $\overline{\nse}$ of bidirected-learned GCNII
for different window sizes and lead times.
Results generally improve with larger window size and smaller lead time.
}}
\label{tab:ablation}
    \newcommand{\error}[2]{\cell{\qty{#1}{}}{\qty{#2}{}}}
    \newcommand{\nash}[2]{\cell{\footnotesize\qty{#1}{\percent}}{\qty{#2}{\percent}}}

	\begin{sc}
		\footnotesize
		\begin{tabular}{ccccccc}
			\toprule
          \multirow{2}{*}{\vspace*{-1mm}window size [$\qty{}{h}$]} & \multicolumn{6}{c}{lead time [$\qty{}{h}$]} \\ \cmidrule(lr){2-7}
			& 1 & 2 & 3 & 6 & 9 & 12 \\
			\midrule
12 & \nash{98.87}{0.05} & \nash{96.28}{0.16} & \nash{92.99}{0.63} & \nash{82.35}{1.94} & \nash{72.51}{2.01} & \nash{63.16}{9.19}  \\ \midrule
24 & \nash{99.05}{0.06} & \nash{96.89}{0.04} & \nash{94.20}{0.52} & \nash{85.59}{1.41} & \nash{75.98}{2.84} & \nash{67.26}{5.12}  \\ \midrule
36 & \nash{99.04}{0.04} & \nash{97.03}{0.15} & \nash{94.53}{0.24} & \nash{85.51}{2.84} & \nash{78.54}{3.65} & \nash{70.52}{3.01}  \\ \midrule
48 & \nash{99.03}{0.06} & \nash{97.03}{0.12} & \nash{94.77}{0.20} & \nash{87.57}{1.21} & \nash{79.72}{2.12} & \nash{74.08}{2.62}  \\ \midrule
60 & \nash{98.99}{0.06} & \nash{96.89}{0.13} & \nash{94.61}{0.10} & \nash{86.66}{1.50} & \nash{81.01}{2.50} & \nash{75.82}{2.93}  \\ \midrule
72 & \nash{99.02}{0.03} & \nash{96.97}{0.02} & \nash{94.65}{0.31} & \nash{87.59}{1.12} & \nash{80.25}{1.85} & \nash{75.52}{1.73}  \\
			\bottomrule
		\end{tabular}
	\end{sc}
\end{table}

\end{document}